\documentclass[10pt,twocolumn,letterpaper]{article}

\usepackage{main}     

\definecolor{cvprblue}{rgb}{0.21,0.49,0.74}
\usepackage[pagebackref,breaklinks,colorlinks,allcolors=cvprblue]{hyperref}

\usepackage{amsmath}
\usepackage{xcolor}
\usepackage{booktabs}  
\usepackage{multirow} 
\usepackage{graphicx} 
\usepackage[table]{xcolor} 
\usepackage{arydshln}    
\usepackage{algorithm}
\usepackage{algpseudocode}
\usepackage{times}
\usepackage{latexsym}
\usepackage{amsmath}
\usepackage{amssymb}
\usepackage{booktabs} 
\usepackage{enumitem}
\usepackage{graphicx}
\usepackage{pifont}
\usepackage{color}
\usepackage{titletoc}
\usepackage{colortbl}
\usepackage{wrapfig}  
\usepackage{lipsum} 
\usepackage{multirow}
\usepackage{listings}
\usepackage{array}
\usepackage{times}
\usepackage{latexsym}
\definecolor{cvprblue}{rgb}{0.21,0.49,0.74}
\hypersetup{
    pagebackref=true,
    breaklinks=true,
    colorlinks=true,
    allcolors=cvprblue
}
\usepackage{amsmath}
\usepackage{xcolor}
\usepackage{booktabs}  
\usepackage{multirow} 
\usepackage{graphicx} 
\usepackage[table]{xcolor} 
\usepackage{xspace}
\usepackage{arydshln}    
\usepackage{algorithm}
\usepackage{algpseudocode}
\usepackage{times}
\usepackage{latexsym}
\usepackage{amsmath}
\usepackage{amssymb}
\usepackage{booktabs} 
\usepackage{enumitem}
\usepackage{graphicx}
\usepackage{pifont}
\usepackage{color}
\usepackage{titletoc}
\usepackage{colortbl}
\usepackage{wrapfig}  
\usepackage{lipsum} 
\usepackage{multirow}
\usepackage{listings}
\usepackage{array}
\usepackage{listings}
\usepackage{xcolor}
\usepackage{tabularx}  
\usepackage[utf8]{inputenc}
\usepackage{CJKutf8}         
\usepackage[most]{tcolorbox}  
\usepackage{xcolor}         
\usepackage{booktabs}  
\definecolor{myGreen}{RGB}{34, 139, 34}
\definecolor{myRed}{HTML}{FF6347}
\usepackage[T1]{fontenc}
\usepackage[utf8]{inputenc}
\usepackage{microtype}
\usepackage{inconsolata}
\usepackage{graphicx}

\definecolor{myGreen}{RGB}{34, 139, 34}
\definecolor{myRed}{HTML}{FF6347}

\newcommand{\cmark}{\textcolor{myGreen}{\ding{51}}}
\newcommand{\xmark}{\textcolor{myRed}{\ding{55}}}
\newcommand{\titledmodelname}{\includegraphics[width=1.35em]{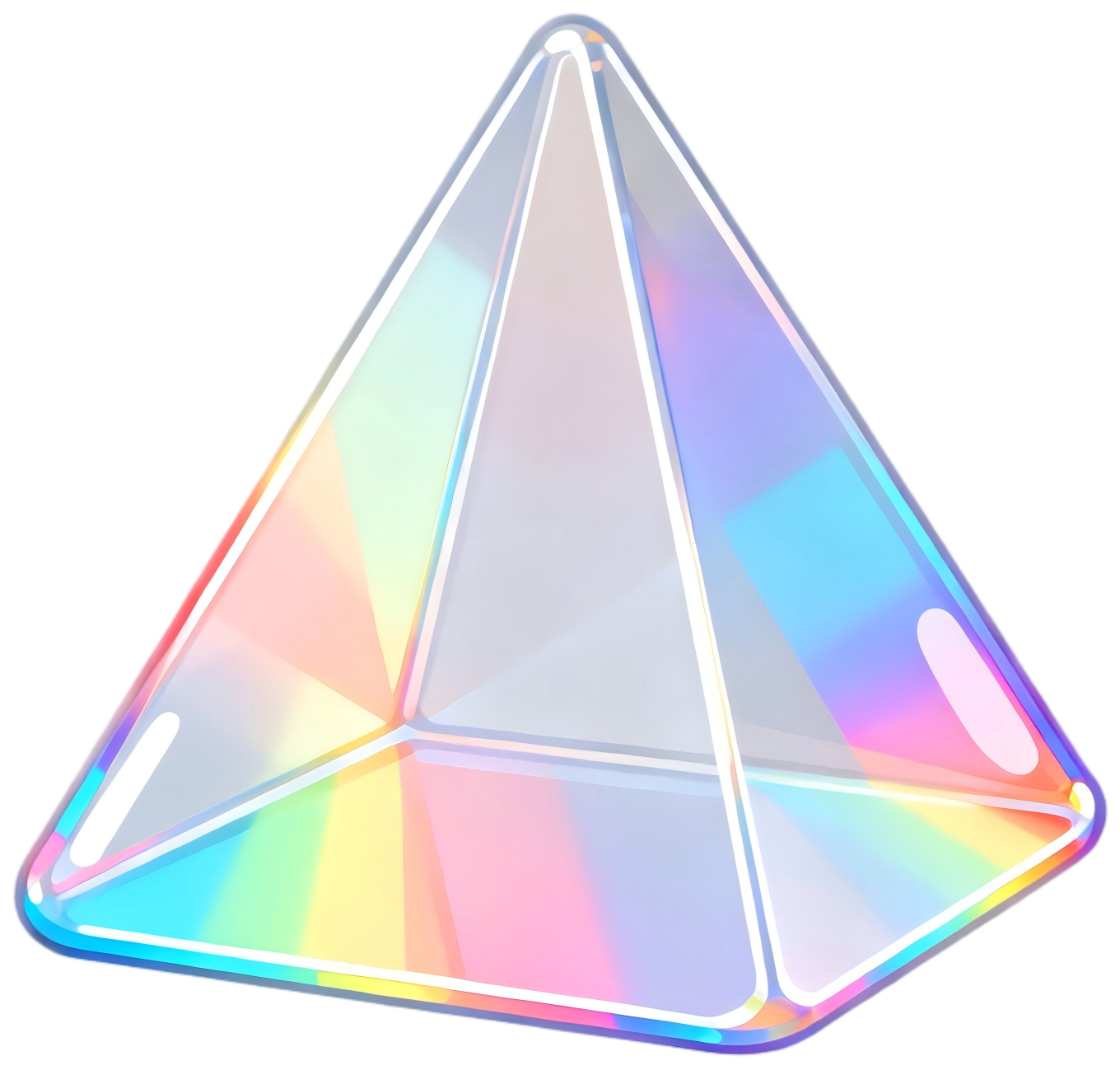}\xspace}

\def\paperID{19861} 
\def\confName{CVPR}
\def\confYear{2026}

\title{PRISM \titledmodelname of Opinions:  A Persona‑Reasoned Multimodal Framework for  User-centric Conversational Stance Detection}

\author{
Bingbing Wang$^{1,4\ast}$, 
Zhixin Bai$^{1}$\thanks{\quad The first two authors contribute equally to this work.}, 
Zhengda Jin$^{1}$,
Zihan Wang$^{1}$,
Xintong Song$^{1}$, 
Jingjie Lin$^{1}$, \\
Sixuan Li$^{5}$,
Jing Li$^{4}$,
Ruifeng Xu$^{1, 2, 3}$\thanks{\quad Corresponding Author} \\
    $^{1}$ Harbin Institute of Technology, Shenzhen, China~
    $^{2}$ Peng Cheng Laboratory, Shenzhen, China \\
    $^{3}$ Guangdong Provincial Key Laboratory of Novel Security Intelligence Technologies \\
    $^{4}$ The Hong Kong Polytechnic University, Hong Kong, China \\
    $^{5}$ Macau University of Science and Technology, Hong Kong, China 
}

\begin{document}
\maketitle
\begin{abstract}
The rapid proliferation of multimodal social media content has driven research in Multimodal Conversational Stance Detection (MCSD), which aims to interpret users’ attitudes toward specific targets within complex discussions. 
However, existing studies remain limited by: 
\textbf{1) pseudo-multimodality}, where visual cues appear only in source posts while comments are treated as text-only, misaligning with real-world multimodal interactions;
and \textbf{2) user homogeneity}, where diverse users are treated uniformly, neglecting personal traits that shape stance expression.
To address these issues, we introduce \textbf{U-MStance}, the first user‑centric MCSD dataset, containing over 40k annotated comments across six real-world targets. 
We further propose \textbf{PRISM}, a \textbf{P}ersona-\textbf{R}easoned mult\textbf{I}modal \textbf{S}tance \textbf{M}odel for MCSD.
PRISM first derives longitudinal user personas from historical posts and comments to capture individual traits, then aligns textual and visual cues within conversational context via Chain‑of‑Thought to bridge semantic and pragmatic gaps across modalities.
Finally, a mutual task reinforcement mechanism is employed to jointly optimize stance detection and stance-aware response generation for bidirectional knowledge transfer.
Experiments on U-MStance demonstrate that PRISM yields significant gains over strong baselines, underscoring the effectiveness of user-centric and context-grounded multimodal reasoning for realistic stance understanding.

\end{abstract}

\section{Introduction}

Social media have become a dominant platform for the expression of opinions, making their analysis crucial to understand divisive societal topics such as politics, public health, and social justice \cite{26-glandt2021stance,27-liang2022zero}. Stance detection, the task of identifying a user’s position toward a target, has become vital for charting public discourse and polarization. As online interactions grow more complex, the field has evolved from simple textual analysis to modeling rich, multi-party conversational contexts \cite{28-li2023improved,29-li2023contextual,30-niu2024challenge}.

\begin{figure}[!t]
  \centering
\includegraphics[width=\linewidth]{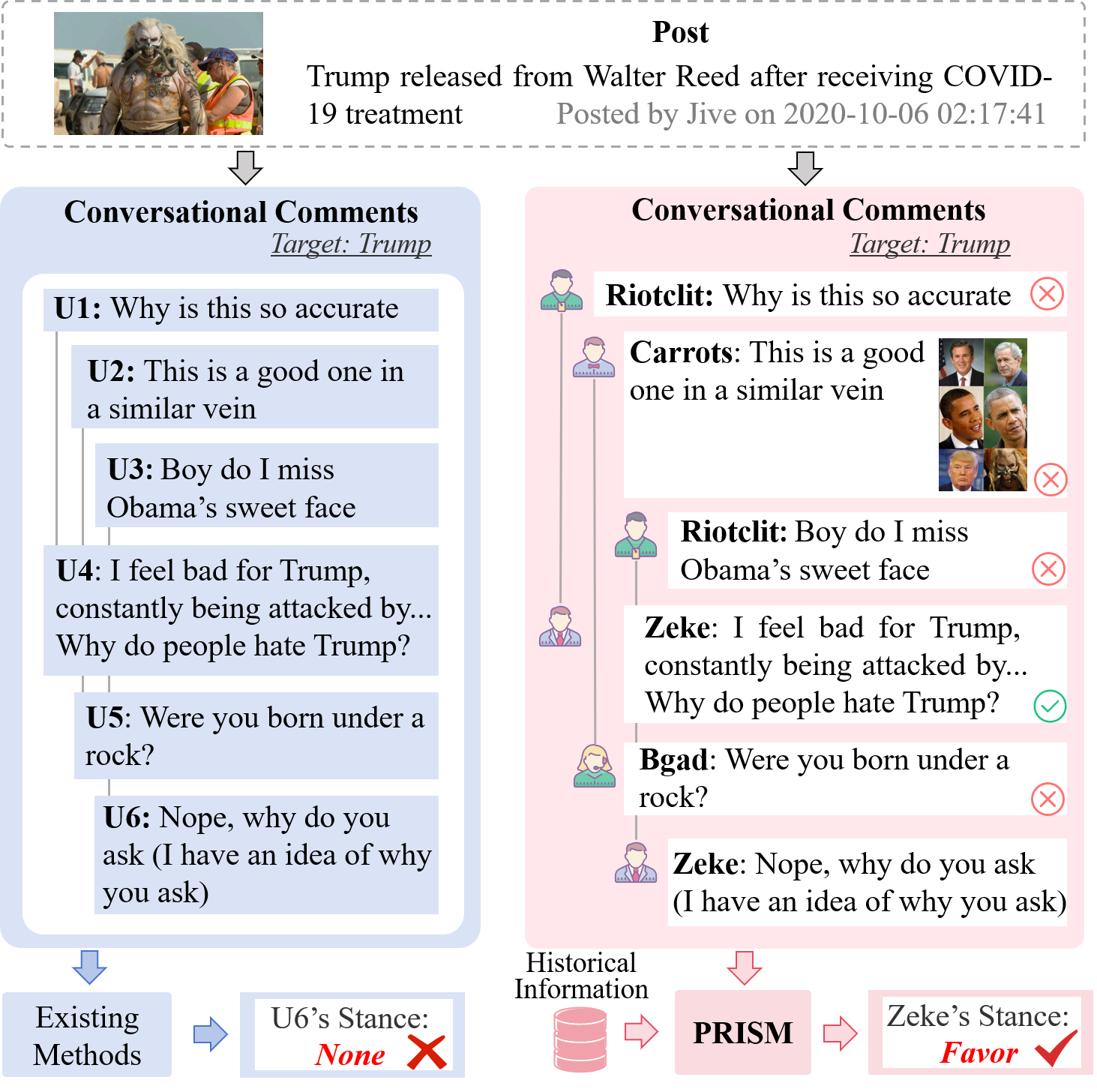}
  \caption{Comparison of a conventional approach for MCSD (left) with our proposed PRISM framework (right).}
  \label{fig_intro}
\end{figure}

The recent shift towards Multimodal Conversational Stance Detection (MCSD), exemplified by the MmMtCSD dataset \cite{1-niu2024multimodal}, marks a crucial step forward. However, despite this progress, two fundamental limitations persist across the field, hindering authentic stance modeling. The first is a \textbf{pseudo-multimodality}, where visual cues appear only in source posts while comments remain text‑only, diverging from the multimodal nature of real-world interactions. The second is \textbf{user homogeneity}, where diverse individuals are modeled uniformly, ignoring personal traits that shape stance expression.
These limitations can be clearly observed in the example shown in Figure \ref{fig_intro}. User \textit{Carrots} combines an image and textual content to contrast Obama and Trump, conveying a derogatory attitude toward Trump, which highlights the necessity of multimodal cues for accurate stance understanding. 
Meanwhile, \textit{Zeke} expresses sympathy for Trump, and \textit{Bgad} challenges this view, revealing conflicting stances. A model that remains blind to their individual personas would fail to grasp the reason for their disagreement, leading to incorrect stance interpretations.

To catalyze progress beyond these limitations, we therefore introduce \textbf{U-MStance}, the first dataset designed for user-centric multimodal conversational stance detection.
Comprising over 40k annotated comments across six real-world targets, U-MStance uniquely incorporates multimodality in both source posts and conversational replies, providing a realistic and challenging benchmark that captures the nuances of user-centric, multimodal interactions.
Building upon this critical new dataset, we further propose \textbf{PRISM}, a \textbf{P}ersona-\textbf{R}easoned mult\textbf{I}modal \textbf{S}tance \textbf{M}odel for MCSD.
Specifically, PRISM first explores the underlying personal tendencies behind stance expressions and derives longitudinal user persona traits inspired by Big Five (OCEAN) theory \cite{44-jiang2024personallm} from historical posts and comments.
Then, it introduces a rationalized cross-modal grounding module, leveraging Chain-of-Thought (CoT) reasoning to explicitly bridge the profound semantic and pragmatic gap between modalities in conversational contexts.
Ultimately, these personas and contextual representations are fed into a mutual task reinforcement mechanism, which jointly optimizes stance classification and stance-aware response generation, enabling bidirectional knowledge transfer that enhances both detection performance and the modeling of user communication styles.
Comprehensive experiments on our U-MStance dataset show that PRISM achieves significant improvements over strong baselines across multiple metrics.
The main contributions can be summarized as follows:
\begin{itemize}
    \item To our knowledge, \textbf{U-MStance} is the first user-centric multimodal conversational stance detection dataset, which extends multimodality beyond posts to comments, and incorporates user information for more realistic and challenging stance understanding.
    \item We propose \textbf{PRISM}, a framework that models user personas and deepens cross-modal correlations for improved user-centric stance detection.
    \item Extensive experiments conducted on our U-MStance dataset demonstrate that PRISM significantly outperforms strong baselines, confirming both its effectiveness and generalization ability.
\end{itemize}

\section{Related Works}
\subsection{Stance Datasets}
The development of stance detection datasets has evolved from analyzing isolated texts to modeling complex interactions.
Early benchmarks like SemEval-2016 Task 6 \cite{2-mohammad2016semeval} and VAST \cite{3-allaway2020zero} focused on sentence-level textual stance detection, establishing foundational models but overlooking the context of dialogue. This limitation spurred the creation of conversational datasets such as SRQ \cite{4-villa2020stance} and CANT-CSD \cite{5-li2023improved}, which treated stance as a dynamic process unfolding within a conversation. 
The most recent advancement is the shift toward multimodal stance detection, which incorporates visual elements alongside text to better reflect modern social media discourse.
However, even the latest datasets for MCSD, such as MmMtCSD \cite{1-niu2024multimodal}, still face two key shortcomings that hinder realistic stance modeling.
First, multimodality is constrained to source posts as static background context, while conversational replies are restricted to text-only.
In addition, information about users, including their posting history, is largely ignored, overlooking contextual factors that shape an individual’s expressed stance and leaving a substantial gap in building comprehensive and authentic stance detection models.

\subsection{Stance Detection Methods}
Methodological advancements in stance detection has spanned multiple task settings in-target \cite{6-li2021multi}, cross-target \cite{7-ding2024cross, 8-liang2021target, 9-wei2019modeling}, and zero-shot scenarios \cite{10-allaway2020zero, 11-liang2022zero, 12-zhu2022enhancing}. Early deep learning approaches primarily fine-tuned pre-trained models such as BERT \cite{13-cambria2018senticnet}. However, the field was revolutionized by the emergence of Large Language Models (LLMs), whose powerful semantic understanding and adaptability introduced new capabilities \cite{14-cheng2024emotion, 15-cheng2024mips, 16-li2024mitigating, 18-ding2024edda}.

Following this paradigm shift, LLM-centric research has moved beyond direct prompting or CoT reasoning toward increasingly sophisticated methodologies \cite{21-zhang2023investigating, 22-zhang2022would}. These include fine-tuning with external knowledge \cite{19-li2023stance}, prompt-based multi-perspective reasoning \cite{7-ding2024cross, 17-wang2025more}, formalizing logic structures \cite{24-zhang2024knowledge}, and designing multi-agent collaborative systems \cite{20-lan2024stance, 25-wang2024deem}. 
Despite this rapid progress, current approaches often treat users as homogeneous, overlooking the personal traits and discourse tendencies that shape stance expression. This limits the interpretation of personalized reactions and conflicting views, necessitating a shift from content-centered to user-centric modeling.

\begin{figure}[!t]
  \centering
  \includegraphics[width=\linewidth]{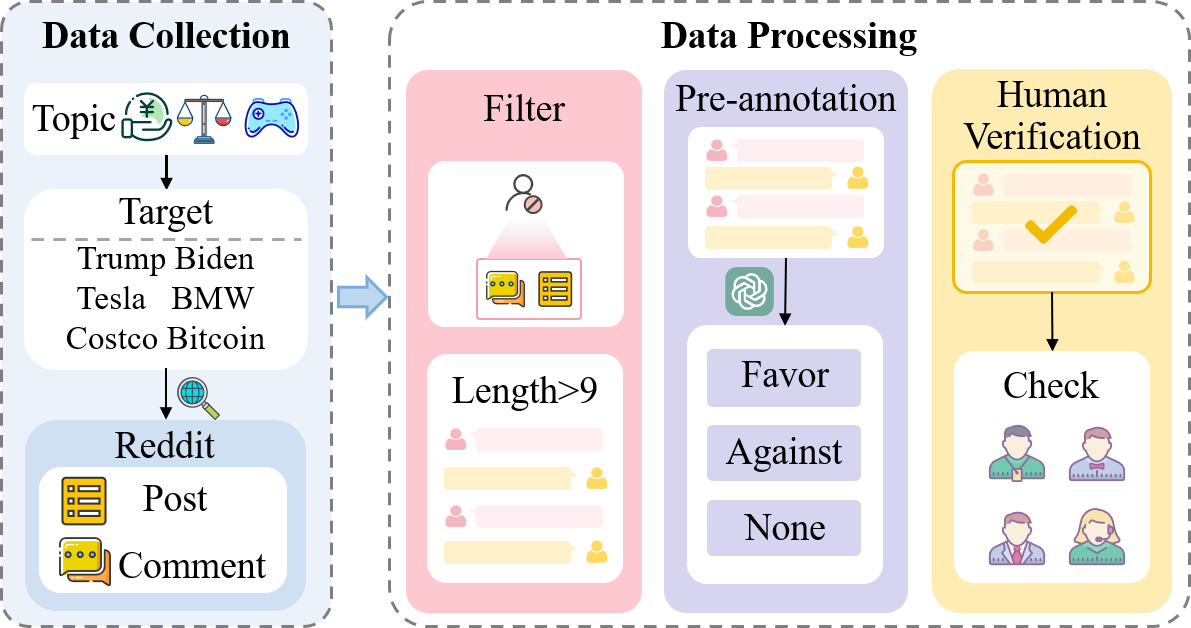}
  \caption{ Data construction pipeline of U-MStance.}
  \label{fig_dataset_overflow}
\end{figure}

\section{Dataset}
This section outlines the construction and key attributes of our U-MStance dataset.
As illustrated in Figure \ref{fig_dataset_overflow}, our data construction pipeline includes collection, processing and verification to ensure data quality and relevance. 
Subsequently, data analysis is conducted to profile the dataset's composition and validate its quality.

\subsection{Data Construction}
\textbf{Dataset Collection}. 
To capture realistic user interactions, we sourced conversational data from Reddit\footnote{https://www.reddit.com}, a large and diverse online forum, using the PRAW library.
Covering the period from 2017 to 2025, we initially identified 835 posts related to six high-engagement, real-world targets: \textit{Trump}, \textit{Biden}, \textit{Tesla}, \textit{BMW}, \textit{Costco}, and \textit{Bitcoin}.
These posts were selected based on popularity metrics like upvotes and comment counts to ensure a sufficient volume of conversational replies.
We also collected each user’s historical posts and comments to enrich their personal information.

\textbf{Filter}.
The raw data underwent systematic processing to ensure quality and consistency. The process began with a filtering stage that removed content from suspended or deleted user accounts and discarded conversational threads with a depth greater than 9 to maintain relevance. 

\textbf{Pre-annotation}.
Following this, GPT-4o-mini was employed for automated pre-annotation, assigning a preliminary stance label (Favor, Against, or None) to each instance relative to its target.

\textbf{Human Verification}.
The automated labels underwent a rigorous human verification process by seven NLP researchers. 
Annotators were first qualified through two pilot rounds and an expert review.
During the main annotation, each instance was independently reviewed by at least two annotators. Disagreements were resolved by a third senior annotator, and the final labels were determined through majority voting. Beyond stance correction, annotators also evaluated the contextual relevance of any associated images. This multi-annotator workflow ensures high label fidelity, and reinforces the overall reliability of the dataset.

\begin{table}[t]
    \caption{Comparison of U-MStance with existing datasets. 
    \textbf{Size}: the number of annotated instances; \textbf{Target}: the number of the targets; \textbf{Conv.}: support for the conversational context; \textbf{Vision-Related}: percentage of vision-integrated instances; \textbf{User}: support for user information; -: not applicable.} 
    \centering
    \footnotesize
    \setlength{\tabcolsep}{3pt}
    \vspace{-1em}
    \resizebox{\linewidth}{!}{
    \begin{tabular}{l|ccccccc}
    \toprule
    \multirow{1}{*}{\textbf{Dataset}} & \multirow{1}{*}{\textbf{Size}} & \multirow{1}{*}{\textbf{Target}} &\textbf{Conv.}& \multirow{1}{*}{\textbf{Vision-Related}} &  \multirow{1}{*}{\textbf{User}}   \\\midrule
       SEM16~\citep{2-mohammad2016semeval} &4,870& 6&\xmark&  \xmark & \xmark \\
       P-stance~\citep{31-li2021p} & 21,574 & 3 & \xmark  &  \xmark & \xmark  \\
       COVID-19~\citep{26-glandt2021stance} & 6,133 & 4 & \xmark &  \xmark & \xmark \\
       WT-WT~\citep{32-conforti2020will} &  51,284 &5 & \xmark &  \xmark & \xmark \\
       VAST~\citep{3-allaway2020zero} & 23,525 & - & \xmark & \xmark  & \xmark\\ 
       SRQ~\citep{4-villa2020stance} &1,348 & 4 & \xmark&  \xmark & \xmark \\
       CANT-CSD~\citep{5-li2023improved} & 5,376 &1 &\cmark& \xmark & \xmark\\
       CTSDT~\citep{29-li2023contextual} &53,861 & 1 &\cmark&  \xmark & \xmark  \\
       MT-CSD~\citep{30-niu2024challenge} &15,876 & 5 & \cmark&\xmark  &\xmark \\
       C-MTCSD~\citep{34-niu2025c}&24,264 & 5 & \cmark & \xmark &\xmark\\
       MT$^{2}$-CSD~\citep{35-niu2025mt2} &24,457 & 6 &\cmark & \xmark & \xmark  \\
       MMVax~\citep{37-weinzierl2023identification} &11,300 & - &\xmark& 100\% &\xmark \\ 
       MMSD~\citep{36-liang2024multi}& 17,544 & 5 &\xmark& 100\% &\xmark \\ 
       MmMtCSD~\citep{1-niu2024multimodal}& 21,340 & 3 &  \cmark & 65.99\% & \xmark \\
       
\midrule
       \textbf{U-MStance(Ours)} & 40,003 & 6 & \cmark &  100\% & \cmark \\
    \bottomrule
    \end{tabular}
    }
    \vspace{-0.3cm}
    \label{tab:data_compare}
\end{table}

\begin{figure*}[!t]
  \centering
  \includegraphics[width=\linewidth]{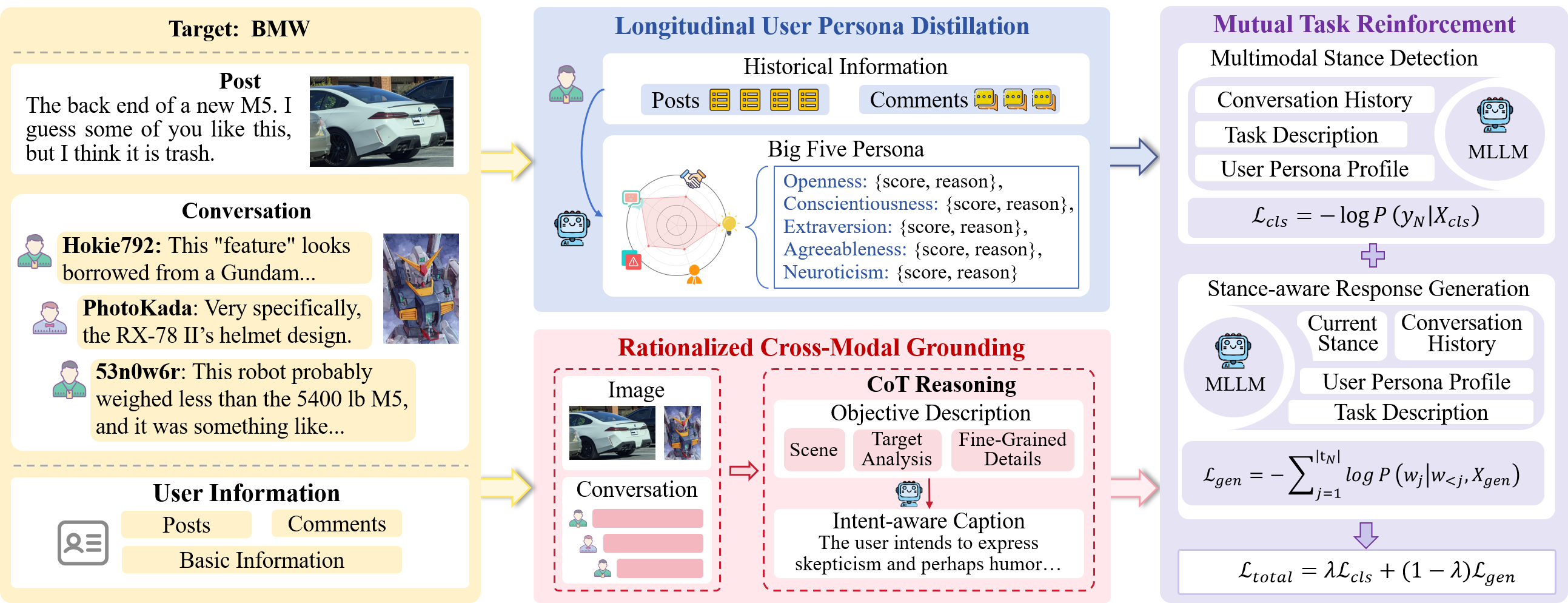}
  \caption{Overview of our PRISM framework.}
  \label{fig_overflow}
  \vspace{-0.5cm}
\end{figure*}

\subsection{Data Analysis}

\textbf{Data Statistics.}
Our dataset comprises 40,003 annotated instances and 24952 user profiles. Each instance consists of a complete conversational thread accompanied by its associated images within the conversation. Crucially, Table \ref{tab:data_compare} provide a comparison between U-MStance and existing datasets. U-MStance is the first dataset in this domain to systematically incorporate user information, offering a foundational resource to address the prevalent issue of user homogeneity. 
More comprehensive statistics, including the distributions of labels and conversational depths, are detailed in Appendix \ref{sec:appendix-Dataset}. 
For model development, the dataset is partitioned into training, validation, and test sets using an 8:1:1 ratio. We strictly partition the data by cross-referencing timestamps to ensure that the three sets are mutually independent.

\textbf{Quality Assessment.} 
To ensure the annotation reliability, we measured inter-annotator agreement using Cohen's Kappa \cite{38-cohen1960coefficient}, achieving an average score of 0.64, indicative of substantial agreement.
This consistency was further reinforced through a rigorous post-annotation review, in which senior researchers audited a randomly selected subset of the annotated data to verify accuracy and resolve borderline cases, providing an additional layer of quality assurance.

\section{Our PRISM Approach}
This section presents a detailed description of our proposed PRISM framework. Given a conversational thread about a global target $T_g$, initiated by a post $P=(t_0,v_0)$, where $t_0$ and $v_0$ are the initial text and image. The post is followed by a sequence of comments $C=\{c_1,...,c_N\}$, where each comment $c_i$ is a tuple $(u_i,t_i,v_i)$ representing the user, their utterance, and an optional image.
Crucially, the input also includes the multimodal historical activity $H_{u_i}$ for each user $u_i$ involved in the conversation, comprising their past posts and comments. Our objective is to predict the stance label $y_N$ of the final comment $c_N$ with respect to $T_g$.

As illustrated in Figure \ref{fig_overflow}, PRISM comprises three key components: 1) \textbf{Longitudinal User Persona Distillation} infers each user’s Big Five-based personality representation from historical posts and comments; 2) \textbf{Rationalized Cross-Modal Grounding} aligns multimodal cues with conversational context using CoT reasoning via an MLLM to generate intent‑aware captions; 3) \textbf{Mutual Task Reinforcement} jointly optimizes stance detection and stance-aware response generation under a multitask objective.

\subsection{Longitudinal User Persona Distillation}
Stance is an externalized reflection of a user’s underlying personal tendencies. Thus, each user is modeled as an entity whose stance expression is shaped by stable personality traits. 
For this purpose, we adopt the empirically validated Big Five (OCEAN) personality model \cite{44-jiang2024personallm}, which characterizes personality along five fundamental dimensions: Openness, Conscientiousness, Extraversion, Agreeableness, and Neuroticism. To infer these traits for user $u_N$, we first aggregate their complete history $H_{u_N}$, which comprises all historical posts and comments.

The distillation process leverages the full power of the MLLM $\mathcal{M}$  to reason over long-form, raw multimodal data, allowing it to capture the nuances of the user. 
To guide the inference, we design a detailed system instruction, $\mathcal{I}_{\text{ocean}}$. This prompt meticulously defines each of the five personality traits and specifies the output format: a numerical score from 1 to 5 for each trait, representing qualitative levels from very low to very high.
The MLLM then generates a structured persona representation, $p_{u_N}$, composed of five numerical ratings. This process is formally expressed as:
\begin{equation}\small
\begin{aligned}
    p_{u_N} &= \{r^{(O)},r^{(C)},r^{(E)},r^{(A)},r^{(N)}\}_{u_N}\\
    &=\mathcal{M}(\mathcal{I}_{ocean},H_{u_N})
\end{aligned}
\end{equation}
where each rating $r\in \{1,2,3,4,5\}$. The resulting profile $p_{u_N}$  is a concise, structured text string that serves as a crucial, personalized context vector. This profile is explicitly injected into the prompt for the final stance detection task, ensuring the model's prediction is conditioned on the stable identity of the user in question. The prompt template can be found in Table \ref{tab:prompt-Persona} of Appendix \ref{sec:appendix-Prompt}.

\subsection{Rationalized Cross-Modal Grounding}
Images in online conversations are not decorations but communicative acts whose pragmatic meaning is shaped by the conversational context.
To decipher this context-dependent rhetorical intent, we propose the Rationalized Cross-Modal Grounding (RCMG) module to generate an intent-aware caption for each image in the conversation.
Inspired by Chain-of-Thought \cite{45-wei2022chain}, this module employs a two-stage reasoning process to bridge the semantic-pragmatic gap. 
The complete prompt template can be found in Table \ref{tab:prompt-Intent} of Appendix \ref{sec:appendix-Prompt}.

\textbf{Objective Description.}
The MLLM $\mathcal{M}$ first performs a context-free analysis of the image $v_i$, producing a factual, objective description $x_i^{obj}$ that grounds reasoning in the literal visual content. This step is guided by a descriptive instruction $I_{desc}$:
\begin{equation}
    x^{obj}_i = \mathcal{M}(I_{desc},v_i)
\end{equation}

\textbf{Intent-aware Interpretation.}
The objective description $x_i^{obj}$ serves as an intermediate reasoning step to support the more complex task of intent inference. At this stage, the MLLM is prompted to infer the speaker's underlying motivation. We construct a prompt with a specific instruction $\mathcal{I}_{int}$, the image $v_i$, its objective description $x_i^{obj}$, and the textual content of the current conversation thread $C_{text}$. The model then generates the final intent-aware caption $\hat x_i$:
\begin{equation}
\hat{x}_i = \mathcal{M}(I_{int}, v_i, x_i^{obj}, C_{text})
\end{equation}
The resulting caption $\hat{x}_i$ captures the rhetorical function of the image within the conversation.

\subsection{Mutual Task Reinforcement}
To cultivate a comprehensive, user-centric understanding of stance, PRISM employs a multitask learning paradigm that jointly optimizes two inherently correlated tasks.
The core intuition is that by training the model to generate a user's response as an auxiliary task, it derives a deeper understanding of the pragmatic and user-specific cues essential for the primary task of stance prediction. This mutual reinforcement serves as a robust regularizer, enhancing the stability and prediction accuracy.

\textbf{Stance Detection (Primary Task).}
The main objective is to predict the stance label $y_N$ of the final comment $c_N$. 
To accomplish this, the MLLM is provided with a comprehensive set of conditioning information including the global target $T_g$, the persona representation of the final user $p_{u_N}$, the multimodal conversational comment $C$, and the set of all intent-aware captions $\mathcal{X}$ generated by the RCMG module for every image in the conversation. 
This combination of raw multimodal data and explicit textual reasoning allows the model to integrate these complementary signals. Formally, we denote the full conditioning set as:
\begin{equation}
    X_{cls}=\{T_g,p_{u_N},C,\mathcal{X}\}
\end{equation}

The classification loss is the negative log-probability of the ground-truth label $y_N$.
\begin{equation}
\begin{aligned}
    \mathcal{L}_{cls} = -\log P(y_N|X_{cls})
\end{aligned}
\end{equation}

\textbf{Stance-aware Response Generation (Auxiliary Task).}
The auxiliary task involves generating the text utterance of the final comment $c_N$.
This task is conditioned on all preceding context, including the original multimodal conversational comments up to the penultimate turn $C_{<N}=\{c_1,...,c_{N-1}\}$ and the corresponding set of intent-aware captions $\mathcal{X}_{<N}$ for all images appearing in the preceding turns. The conditioning information also includes the final user's persona $p_{u_N}$ and their ground-truth stance $y_N$. We denote this conditioning set as:
\begin{equation}
    X_{gen}=\{T_g, y_N, p_{u_N}, C_{<N}, \mathcal{X}_{<N}\}
\end{equation}

The generation loss is the standard negative log-likelihood over the tokens $w_j$ of the utterance $t_N$:
 \begin{equation}
\begin{aligned}
    \mathcal{L}_{gen} = - \sum_{j=1}^{|t_N|} \log P(w_j | w_{<j},X_{gen}) 
\end{aligned}
\end{equation}

\textbf{Joint Optimization.}
The final training objective is a weighted sum of the two losses:
\begin{equation}
\mathcal{L}_{total} = \lambda \mathcal{L}_{cls} + (1-\lambda) \mathcal{L}_{gen}
\end{equation}

This mutual reinforcement ensures PRISM learns a holistic model of stance expression.

\begin{table*}[t!]
\caption{In-target main results (\%), where the methods based on LLMs involve testing through direct questioning. Best and second-best scores are in \textbf{bold} and \underline{underline}, respectively. $^*$ indicates our PRISM model significantly outperforms baselines ($p < 0.05$). The dashed line separates fine-tuned from non-fine-tuned methods. w/ denotes the backbone. Ag, Fa, Avg represent F1-against, F1-favor, and F1-avg.}
\vspace{-0.2cm}
\centering
\small
\setlength{\tabcolsep}{0.5pt}
\setlength{\extrarowheight}{0.5pt}
\renewcommand{\arraystretch}{1.1}
\resizebox{\textwidth}{!}{%
\begin{tabular}{ll cccccccccccccccccccc}
\toprule
\multirow{2.5}{*}{Modality} & \multirow{2.5}{*}{Method} & \multicolumn{3}{c}{Trump} & \multicolumn{3}{c}{Biden} & \multicolumn{3}{c}{Tesla} & \multicolumn{3}{c}{BMW}& \multicolumn{3}{c}{Costco}& \multicolumn{3}{c}{Bitcoin} &\multirow{2.5}{*}{Overall} \\
\cmidrule(lr){3-5} \cmidrule(lr){6-8} \cmidrule(lr){9-11} \cmidrule(lr){12-14} \cmidrule(lr){15-17}\cmidrule(lr){18-20}  
& & Ag & Fa & Avg &Ag & Fa & Avg  & Ag & Fa & Avg &Ag & Fa & Avg &Ag & Fa & Avg &Ag & Fa & Avg  \\
\midrule
\multirow{6}{*}{Text-only} & 
BERT 
&65.94	&11.57&	38.75	&65.31	&65.31	&54.46	&56.07	&39.61	&47.85&	64.22&	53.66&	58.94	&51.83	&41.94&	46.89&	36.96&	63.43	&50.19& 49.51\\
& RoBERTa 
&66.81	&24.19	&45.51&	64.37&	64.37	&58.39&	57.59&	47.62&	52.61	&64.45	&54.07	&59.26&	62.14&	54.94&	58.54&	49.24&	68.89&	59.07&55.56  \\
& PoliBERTweet 
& 60.00& 	30.00&	45.00&	58.70&	57.58&	58.14&	41.67&	34.69	&38.18&	60.00	&54.28&	57.14&	57.56&	51.41&	54.48&	28.07&	58.97&	43.52&49.41 \\
\cdashline{2-21}
& LLaMA2 
& 24.32&	10.26&	17.29&	23.81	&17.50&	20.65&	32.62&	32.50&	32.56&	35.74&	33.45&	34.60&	30.73&	38.76&	34.75&	26.60&	42.74&	34.67&29.09\\
& GPT-3.5
&62.27	&27.59&	44.93&	51.14&	58.80&	54.97&	63.92&	45.18	&54.55	&60.73&	50.22&	55.47&	62.03&	52.31&	57.17&	47.97&	65.44&	56.71&53.97 \\
& GPT-4
&67.52&	38.71&	53.11&	56.28&	67.34&	61.81&	66.34&	55.19&	60.77&	67.76&	61.32&	64.54&	63.35&	60.95&	62.15&	56.44&	67.73&	62.08&60.74 \\
\midrule
\multirow{5}{*}{Multimodal} 
& BERT+ViT 
&63.98&	25.85&	44.91&	49.33&	61.45&	55.39&	56.46&	40.27&	48.37&	65.00&	55.56&	60.28&	62.84&	51.90&	57.37&	34.67&	67.37&	51.02&52.89 \\
& TMPT 
&66.02&	27.91&	46.97&	50.69&	62.07&	56.38&	53.08&	34.97&	44.02&	61.30&	58.38&	59.84&	60.68&	52.39&	56.54&	44.07&	66.31&	55.18 &53.16 \\

\cdashline{2-21}
& Qwen-VL 
& 62.50&	24.33&	43.42&	49.49&	54.63&	52.06&	50.44&	44.23&	47.34&	53.87&	42.64&	48.26&	62.60&	54.82&	58.71&	50.45&	69.60&	60.03&51.64\\
& LLaVa 
& 53.18&	19.94&	36.56&	30.23&	46.15	&38.19	&43.09&	31.95&	37.52&	40.30&	26.89&	33.60&	44.92	&31.39	&38.16&	22.88&	55.13&	39.00&	37.17\\
& MiMo 
& 59.08&	21.05&	40.07	&40.63&	48.84&	44.73&	49.53&	37.91	&43.72	&52.60&	30.65&	41.62&	54.76&	38.06&	46.41&	39.02&	57.22&	48.12&	44.11\\
& GPT4-1
&69.94&	47.74&	58.84&	55.49&	66.48&	60.99&	\underline{74.14}&	\underline{61.37}&	67.75&	70.20&	\underline{70.52}&	70.36&	\textbf{74.88}&	\textbf{72.40}&	\textbf{73.64}&	58.94&	72.82&	65.88 &66.24\\

\midrule
\rowcolor{gray!20} 
\multicolumn{2}{c}{\textbf{PRISM (ours)}} &\underline{73.84}&	\underline{48.05}&	\underline{60.94} &	\underline{58.82}&	\textbf{73.22}$^*$&	\textbf{66.02}$^*$&	72.06&	\textbf{63.64}$^*$&	\underline{67.85}&	\underline{75.00}&	67.61&	\underline{71.30}&	\underline{72.44}&	65.72&	69.08&	\underline{72.00}&	\underline{79.45}&	\underline{75.72}&\underline{68.49}\\
\multirow{2}{*}{Backbones}
&- w/ LLaVA & 70.47&	37.42&	53.94&	40.00&	63.00&	51.50&	67.16&	49.79	&58.47&	67.63&	68.44&	68.04&	69.93&	63.60&	66.77&	60.91&	73.47	&67.19&	60.99\\
&- w/ MiMo& \textbf{74.40}$^*$&	\textbf{53.01}$^*$	&\textbf{63.71}$^*$	&\textbf{60.61}$^*$&	\underline{69.96}&	\underline{65.28}&	\textbf{75.12}$^*$&	61.22&	\textbf{68.17}$^*$&	\textbf{75.80}$^*$&	\textbf{76.86}$^*$&	\textbf{76.33}$^*$&	71.72&	\underline{68.36} &	\underline{70.04} &	\textbf{73.22}$^*$&	\textbf{81.53}$^*$&	\textbf{77.38}$^*$&\textbf{70.15}$^*$ \\

\bottomrule
\end{tabular}%
}
\vspace{-0.2cm}
\label{tab:in-target}
\end{table*}

\section{Experiment}

\subsection{Experimental Setup}
\textbf{Implementation details.} 
Our framework is built on Qwen2.5-VL-7B \cite{48-bai2025qwen2} and fine-tuned for 5 epochs using our multitask objective, with the loss-weighting hyperparameter $\lambda$ set to 0.7. We use the AdamW optimizer \cite{51-adam2014method}, with an initial learning rate of $1\times 10^{-5}$, a cosine decay schedule, and a global batch size of 16. All experiments were run on a server with four NVIDIA A100 (40 GB) GPUs using PyTorch and the Hugging Face Transformers library.

\textbf{Evaluation metrics.}
We adopt the average F1 score (F1-avg) as the primary evaluation metric, following \cite{1-niu2024multimodal} and \cite{34-niu2025c}.
F1-avg is computed as the mean of F1 scores for the against and favor stances, denoted as F1-against and F1-favor.

\subsection{Baseline Methods}
We conduct extensive experiments with a range of existing methods, categorized into text-only and multimodal approaches. All baseline models are fine-tuned on our specialized training data, with the exception of the GPT series, which is evaluated in a zero-shot setting. 
\textit{\textbf{Text-only baselines}}. We evaluate the following methods: \textbf{BERT} \cite{39-devlin2019bert}, \textbf{RoBERTa} \cite{40-liu2019roberta}, \textbf{PoliBERTweet} \cite{43-kawintiranon2022polibertweet}, \textbf{LLaMA2} \cite{49-touvron2023llama}, \textbf{GPT-3.5} and \textbf{GPT-4}. 
\textit{\textbf{Multimodal baselines}}. We include: \textbf{BERT-ViT} \cite{36-liang2024multi}, \textbf{TMPT} \cite{36-liang2024multi}, \textbf{LLaVA} \cite{47-liu2024improved}, \textbf{Qwen-VL} \cite{48-bai2025qwen2}, \textbf{MiMo} \cite{46-xiaomi2025mimo} and \textbf{GPT4-1}. For more details, please refer to  Appendix \ref{sec:appendix-Model}.

\begin{table*}[t!]
\caption{Comparison of different models for cross-target stance detection, where the methods based on LLMs involve testing through direct questioning. Best and second-best scores are in \textbf{bold} and \underline{underline}, respectively. $^*$ indicates our PRISM model significantly outperforms baselines ($p < 0.05$). The dashed line separates fine-tuned from non-fine-tuned methods. w/ denotes the backbone. Ag, Fa, Avg represent F1-against, F1-favor, and F1-avg.}
\vspace{-0.2cm}
\centering
\small
\setlength{\tabcolsep}{0.5pt}
\renewcommand{\arraystretch}{1.1}
\resizebox{\textwidth}{!}{%
\begin{tabular}{ll cccccccccccccccccccc}
\toprule
\multirow{2.5}{*}{Modality} & \multirow{2.5}{*}{Method} & \multicolumn{3}{c}{Trump} & \multicolumn{3}{c}{Biden} & \multicolumn{3}{c}{Tesla} & \multicolumn{3}{c}{BMW}& \multicolumn{3}{c}{Costco}& \multicolumn{3}{c}{Bitcoin}&\multirow{2.5}{*}{Overall} \\
\cmidrule(lr){3-5} \cmidrule(lr){6-8} \cmidrule(lr){9-11} \cmidrule(lr){12-14} \cmidrule(lr){15-17}\cmidrule(lr){18-20}  
& & Ag & Fa & Avg &Ag & Fa & Avg  & Ag & Fa & Avg &Ag & Fa & Avg &Ag & Fa & Avg &Ag & Fa & Avg  \\
\midrule
\multirow{3}{*}{Text-only} & 
BERT 
&1.72&	19.76	&10.74&	23.75&	25.18	&24.27	&26.55&	16.21&	21.38&	39.46&	36.69&	34.02	&38.06&	16.35	&28.99&	21.27&	29.30&	25.28&24.11 \\
& RoBERTa 
&14.69&	25.94&	20.32	&36.72&	40.15&	38.43&	35.19&	21.39	&28.29&	53.04&	35.78&	47.15&	2.89&	5.66&	6.44&	35.08&	29.64&	32.36 &28.83 \\
& PoliBERTweet 
& 42.85	&3.84	&23.35&	33.96&	43.64&	38.80&	35.46&	30.93&	33.19&	48.49&	31.03&	39.76	&9.71&	20.00&	14.85&	25.00&	36.36&	30.68&30.11 \\
\midrule
\multirow{2}{*}{Multimodal} 
& BERT+ViT & 6.83&	23.91&	15.37&	31.21&	28.69&	29.94&	23.58&	16.00&	19.79&	44.50&	32.71&	38.60&	7.58&	23.77&	15.68&	27.11&	21.82&	28.19&24.60 \\
& TMPT 
& 7.90&	25.12	&16.52&	31.75&	29.66&	30.75&	13.98&	7.41&	10.69&	47.39&	19.86&	33.62&	11.63&	15.31	&13.47&	30.61&	38.68&	34.65&23.28 \\

\midrule
\rowcolor{gray!20} 
\multicolumn{2}{c}{\textbf{PRISM (ours)}} &\underline{59.50}&	\underline{28.43}&	\underline{43.97}&	\textbf{48.54}$^*$	&\underline{61.59}	&\underline{55.06}&	\underline{62.30}&	\textbf{56.89}$^*$&	\underline{59.60}&	\textbf{66.06}$^*$&	51.02&	58.54	&45.16&	\textbf{57.75}$^*$&	\underline{51.45}&	\textbf{60.50}$^*$&	\underline{67.70}&	\underline{64.10}&\underline{55.45}\\
\multirow{2}{*}{Backbones}
&- w/ LLaVA&53.31	&22.35&	37.83&	35.24	&59.63&	47.43&	59.39&	49.81	&54.60&	58.52&	\textbf{58.82}$^*$	&\underline{58.67}&	\underline{46.63}&	43.26&	44.94&	56.07&	61.02&	58.55	&50.34 \\
&- w/ MiMo&\textbf{60.73}$^*$&	\textbf{34.39}$^*$&	\textbf{47.56}$^*$&	\underline{43.43}&	\textbf{67.03}$^*$&	\textbf{55.23}$^*$&	\textbf{67.90}$^*$&	\underline{56.37}&	\textbf{62.14}$^*$&	\underline{64.47}&	\underline{53.63}&	\textbf{59.05}$^*$&	\textbf{47.47}$^*$&	\underline{56.92}&	\textbf{52.20}$^*$&	\underline{60.26}&	\textbf{68.90}$^*$&	\textbf{64.58}$^*$&\textbf{56.79}$^*$\\
\bottomrule
\end{tabular}
}

\vspace{-0.3cm}
\label{tab:cross-target}
\end{table*}

\subsection{Main Results}
\textbf{In-Target Stance Detection.}
We report the experimental results on the U-MStance dataset under the in-target setup, where the training and testing sets share identical targets. As shown in Table \ref{tab:in-target}, our proposed PRISM framework achieves the highest overall performance with an F1-avg of 68.49\%, which substantially outperforms multimodal GPT4-1 with 66.24\% and the text-only GPT-4 with 60.74\%. This improvement highlights PRISM's effectiveness and efficiency in capturing multimodal stance representations even with a relatively lightweight backbone.
The results also confirm the importance of visual signals for stance understanding, as evidenced by consistent gains from multimodal models such as BERT-ViT compared to BERT, and GPT4-1 compared to GPT-4.
By incorporating rationalized cross-modal grounding, PRISM effectively aligns textual and visual semantics, producing more accurate and context-aware stance predictions.
Interestingly, the large models, such as LLAVA and MiMo, generally underperform compared to smaller, specialized models like TMPT. Despite their strong general language understanding, these larger models may be less effective at capturing fine-grained, conversation-specific, and multimodal cues that are critical for stance detection.
In contrast, our PRISM leverages user persona and multimodal modeling to effectively capture nuanced cues, enabling its LLM backbone to achieve superior performance.

\textbf{Cross-Target Stance Detection.}
We further evaluate generalization under the cross-target setting, where targets in the test set are unseen during training, as summarized in Table \ref{tab:cross-target}.
Specifically, we conduct cross-target evaluations between pairs such as Trump-Biden, Tesla-BMW, and Costco-Bitcoin, training on one and testing on the other. 
This scenario is more challenging, requiring the model to transfer stance understanding across targets with differing semantics and discourse style.
We can observe that most smaller models suffer a notable degradation in performance, revealing their limited capacity to generalize beyond the specific targets encountered during training. 
Remarkably, our proposed PRISM framework maintains a relatively high and stable performance across all cross-target evaluations, demonstrating superior robustness and adaptability. 
This advantage stems from two components: the integration of user persona modeling, which allows the model to capture consistent user-level stance tendencies across topics, and the mutual task reinforcement mechanism, which jointly optimizes stance prediction and rationalized cross-modal grounding. Together, these components enable PRISM to establish a more target-invariant representation of stance, thereby enhancing its generalization capability across heterogeneous contexts.

\subsection{Impact of Different Backbone}
We further assess the adaptability of PRISM by integrating it with different MLLM backbones under both the in-target and cross-target settings, as illustrated in Table \ref{tab:in-target} and Table \ref{tab:cross-target}. 
Specifically, we replace the original Qwen-VL backbone with LLaVA and MiMo to examine how backbone choice affects performance. Across both evaluation setups, MiMo provides consistent performance gains across both evaluation setups.
This finding demonstrates not only the effectiveness but also the generalizability of our framework, showing that PRISM can seamlessly adapt to stronger multimodal backbones while preserving its core advantages. 
The gains from MiMo likely stem from its enhanced multimodal reasoning capabilities and more robust cross-domain representation learning. These properties align well with PRISM’s design, enabling more precise modeling of nuanced stance cues and improving the transferability of stance understanding across heterogeneous targets.

\begin{table}[!t]
\caption{Experimental results (\%) of the ablation study. w/o means without.}
\vspace{-0.2cm}
\centering
\small
\setlength{\tabcolsep}{5pt} 
\renewcommand{\arraystretch}{1.1}
\resizebox{\linewidth}{!}{
\begin{tabular}{lcccccc}
\toprule
 & Trump &Biden& Tesla &BMW &Costco& Bitcoin\\
\midrule
\rowcolor{gray!20}
\multicolumn{7}{c}{\textit{\textbf{In-Target}}} \\ 
PRISM & \textbf{60.94} &\textbf{66.02} & \textbf{67.85}& \textbf{71.30} &\textbf{69.08}&\textbf{75.72}  \\ \hdashline
-w/o Persona & 49.56&60.38&61.67&70.02&64.85&72.92\\
-w/o Intent & 50.81 & 57.30 & 56.45 & 65.34 & 61.60&72.59 \\ 
-w/o Mutual & 59.64&64.34&67.07&68.97&68.89&73.34\\
\hdashline
\rowcolor{gray!20}
\multicolumn{7}{c}{\textit{\textbf{Cross-Target}}} \\ 
PRISM & \textbf{43.97}& \textbf{55.06}& \textbf{59.60} & \textbf{58.54} & \textbf{51.45} &\textbf{64.10} \\  \hdashline
-w/o Persona &40.96&47.16&56.42&55.95&50.13&60.73 \\
-w/o Intent & 40.99 & 45.19 & 41.90& 48.30& 47.84& 53.66\\ 
-w/o Mutual & 43.63 & 52.16 & 58.80& 54.01 & 49.88 & 63.91\\
\bottomrule
\end{tabular}
}
\vspace{-0.5cm}
\label{tab:ablation}
\end{table}

\begin{figure*}[!t]
  \centering
  \includegraphics[width=1\linewidth]{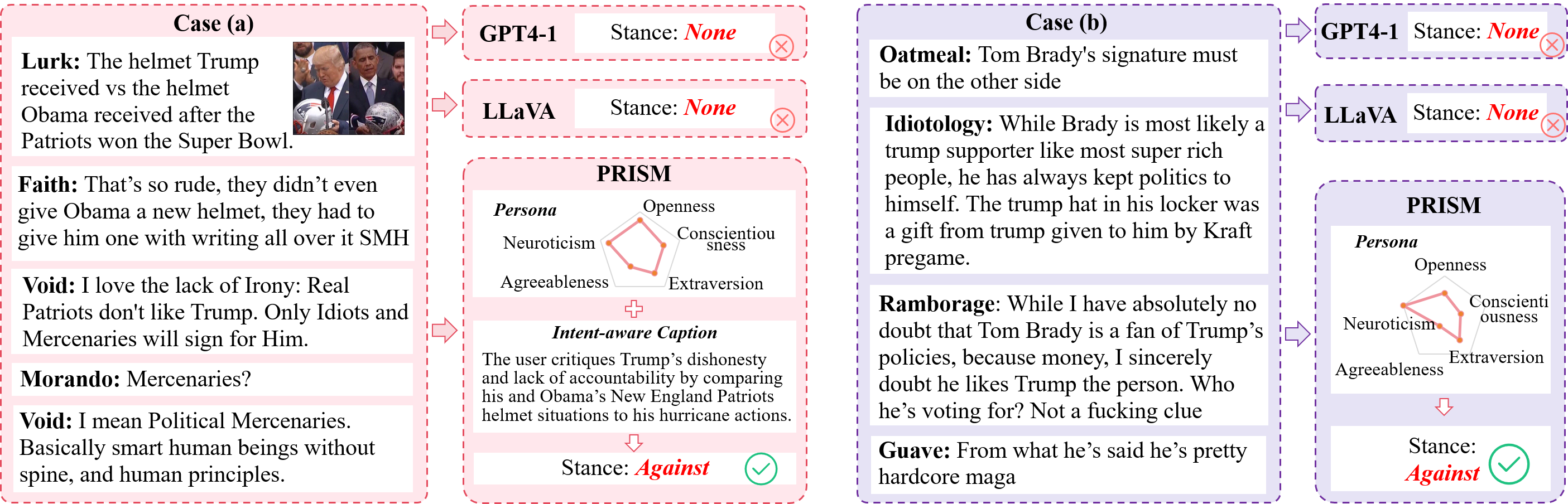}
  \caption{Case study comparing PRISM with baseline models.}
  \label{fig_case}
  \vspace{-0.5cm}
\end{figure*}

\subsection{Ablation Study}
To evaluate the contribution of each component in our PRISM framework, we conduct a comprehensive ablation study with the results presented in Table \ref{tab:ablation}. 
Replacing the persona representation $p_{u_N}$ with the user's original raw information (\textbf{w/o Persona}) consistently drops performance across all targets. 
This underscores the critical role of persona distillation in filtering inherent noise and extracting task-relevant characteristics, enabling PRISM to more precisely interpret individualized, user-centric stance expressions.

Substituting intent-aware captions $\hat x$ with raw visual features (\textbf{w/o Intent}) results in a more substantial degradation, highlighting the crucial role of rationalized cross-modal grounding.
By explicitly aligning the image pragmatic intent with textual context, this component enables the PRISM to correctly infer subtle or implicit visual cues essential for multimodal stance detection.
Lastly, discarding the mutual task reinforcement mechanism and reverting to a single-task stance detection objective (\textbf{w/o Mutual}) also leads to a noticeable decline in performance. 
This confirms that jointly optimizing stance detection with auxiliary generation task encourages beneficial interactions between multimodal understanding and stance reasoning, ultimately producing more robust and transferable representations.

\subsection{Case Study}
To visually demonstrate the superiority of our method and the inherent challenges of the task, Figure \ref{fig_case} provides a case study.
In Case (a), both GPT4-1 and LLaVA fail to detect the sarcastic implication conveyed by joint multimodal cues, incorrectly assigning a neutral stance due to their reliance on surface-level information.
In contrast, our PRISM accurately predicts an \textit{Against} stance.
This improvement stems from its rationalized cross-modal grounding, which captures the user’s implicit criticism expressed through the comparison embedded in the image, as well as its persona-aware interpretation of the user’s communicative tendencies.

Meanwhile, Case (b) presents a more subtle scenario, where sarcasm is intertwined with the user’s disposition.
Again, GPT4-1 and LLaVA misclassify the stance, likely due to their lack of user-level modeling and reliance solely on the current conversational snippet.
PRISM, however, correctly identifies the \textit{Against} stance by leveraging its persona module to interpret longitudinal behavioral patterns. 
Specifically, the persona visualization highlights a high Neuroticism score, reflecting an emotionally expressive and critical style. Incorporating such personality cues enables PRISM to infer the user’s consistent critical orientation even when sarcasm is subtle.

\begin{figure}[!t]
  \centering
  \includegraphics[width=1\linewidth]{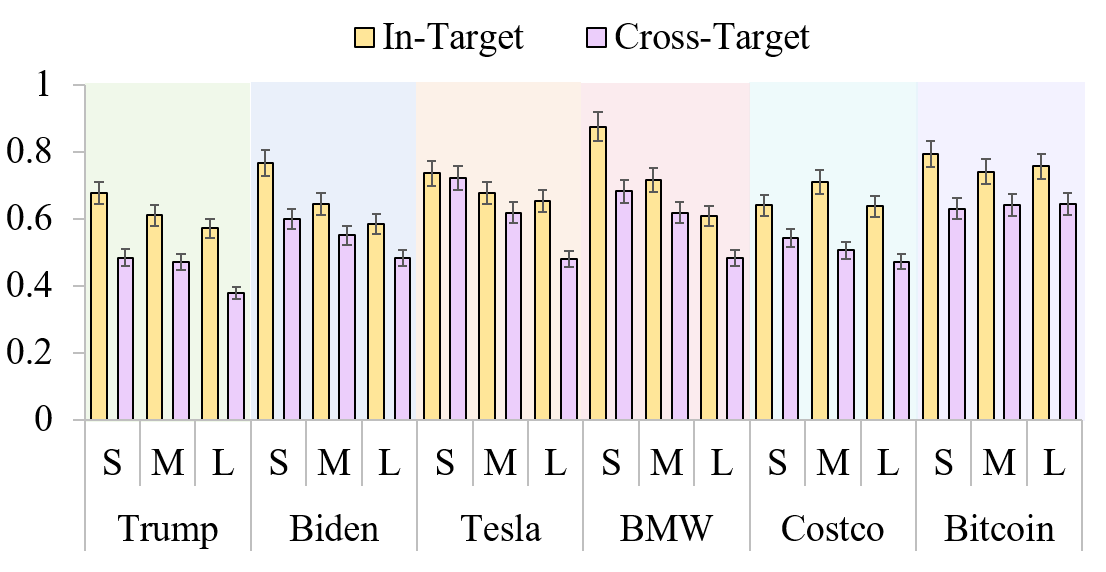}
  \caption{Performance of different depths. Conversational depth: S (1-2 turns), M (3-4 turns), L ($\geq$5 turns).}
  \label{fig_depth}
  \vspace{-0.5cm}
\end{figure}

\subsection{Impact of Conversation Depth}
We further investigate evaluate how PRISM's stance detection performance is influenced by varying conversational depths.
As shown in Figure \ref{fig_depth}, performance shows a gradual decline as depth increases, which is expected since deeper conversations involve more intricate discourse structures, longer-range dependencies, and greater contextual ambiguity, making the final stance harder to infer.
Notably, for the targets \textit{Costco} and \textit{Bitcoin}, PRISM maintains relatively stable performance across depths. This robustness is largely attributed to the rationalized cross-modal grounding module, whose intent-aware image captions help align visual and textual cues and preserve salient multimodal information, thereby mitigating the challenges posed by longer conversational contexts.

\section{Conclusion}
This paper presents two critical limitations in MCSD, including pseudo-multimodality, where visual cues are limited to source posts, and user homogeneity, which neglects individual differences.
To address these, we propose U-MStance, the first user-centric MCSD dataset with multimodal and user information. Furthermore, we design PRISM, a framework that distills stable user personas, grounds image intent, and reinforces stance understanding through multitask learning.
Extensive Experiments on U-MStance demonstrate that PRISM consistently outperforms strong baselines and exhibits robust generalization.

\section*{Limitations}
Despite the significant improvements achieved by our model, certain limitations remain. First, while U-MStance spans diverse domains including politics, automotive, consumer goods, and cryptocurrency, it lacks coverage of highly specialized or professional sectors, such as specific legal statutes or cutting-edge scientific controversies. Second, as illustrated in Figure \ref{fig_depth}, model performance exhibits a declining trend as conversational depth increases. In extremely deep or multi-branch discussion threads, the model still encounters challenges in perfectly capturing subtle logical transitions and long-range contextual drifts.

{
    \small
    \bibliographystyle{ieeenat_fullname}
    \bibliography{main}
}

\appendix

\section{Detailed Dataset Statistics}
\label{sec:appendix-Dataset}
\begin{figure}[!h]
  \centering
  \includegraphics[width=0.7\linewidth]{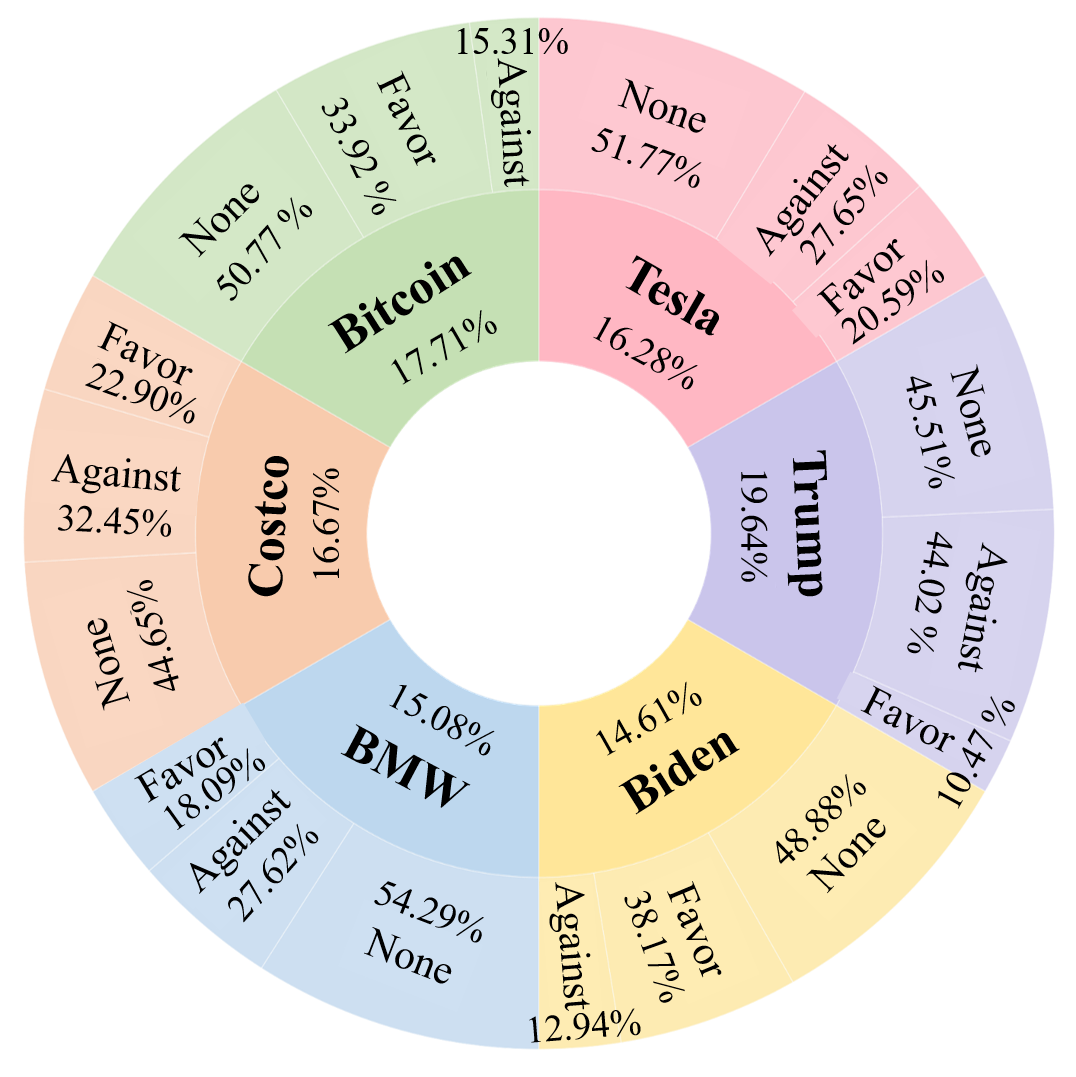}
  \caption{The distribution of stance categories in our U-MStance.}
  \label{fig_dataset_label}
\end{figure}

Figure \ref{fig_dataset_label} illustrates the distribution of instances across different labels.
Table \ref{tab:Appendix} illustrates the distribution of instances across different conversational depths,  including the number of users and images for each target. Within our collected dataset, \textit{36}\% of the posts contain a single image, while \textit{63}\% include two or more images, which means visual information is also present within the comments.
In general, our dataset features more extensive conversational depth compared to existing conversational stance detection datasets, and achieves 100\% vision-related coverage.

\begin{table*}[t] 
\centering
\caption{Comparison of Depth and Interaction Data Across Different Entities}
\label{tab:Appendix}
\small 
\begin{tabularx}{0.85\textwidth}{c c c *{6}{>{\centering\arraybackslash}X} c} 
\toprule
& \textbf{Depth} & \textbf{Avg. WC}  & \textbf{Trump} & \textbf{Biden} & \textbf{Tesla} & \textbf{BMW} & \textbf{Costco} & \textbf{Bitcoin} & \textbf{Total} \\ \midrule
\textbf{User}    & -- & -- & 5087 & 3371 & 3896 & 3956 & 4519 & 4123 & 24952 \\
\textbf{Image}     & -- & -- & 47   & 40   & 55   & 102  & 20   & 21   & 285   \\ 
\cdashline{1-10}
\rowcolor{gray!20}
\textbf{Post}    & 1 & 17.96 & 16   & 27   & 14   & 19   & 8    & 13   & 97 (0.24\%)    \\ 
\multirow{8}{*}{\textbf{Comment}}  & 2 & 27.38 & 927  & 1011 & 723  & 741  & 768  & 1186 & 5356 (13.39\%)   \\
        & 3 & 23.71 & 2306 & 2096 & 1982 & 1816 & 2288 & 2647 & 13135 (32.84\%) \\
        & 4 & 27.65 & 1645 & 1189 & 1402 & 1176 & 1470 & 1292 & 8174 (20.43\%)  \\
        & 5 & 29.99 & 1076 & 694  & 928  & 821  & 1001 & 846  & 5366 (13.41\%)  \\
        & 6 & 31.52 & 795  & 427  & 621  & 637  & 541  & 517  & 3538 (8.84\%)  \\
        & 7 & 35.98 & 529  & 231  & 430  & 385  & 316  & 304  & 2195 (5.49\%)  \\
        & 8 & 39.59 & 333  & 106  & 264  & 290  & 168  & 173  & 1334 (3.33\%)  \\
        & 9 & 45.02 & 231  & 65   & 150  & 148  & 108  & 106  & 808 (2.02\%)   \\ \bottomrule
\end{tabularx}
\end{table*}

\section{Model Details}
\label{sec:appendix-Model}
\begin{table}[H] 
\centering
\caption{Model details including parameters, access type, and their respective sources.}
\label{tab:model_specs}
\small 
\setlength{\tabcolsep}{3pt}
\begin{tabular}{lccc} 
\toprule
\textbf{Model} & \textbf{Parameters} & \textbf{Access} & \textbf{Source} \\ 
\midrule
BERT & <1B & Local & Google \\
RoBERTa & <1B & Local & Meta \\
PoliBERTweet & <1B & Local & Kawintiranon et al. \\
LLaMA-2-7B & 7B & Local & Meta \\
GPT-3.5 & Unknown & API   & OpenAI           \\
GPT-4 & Unknown & API   & OpenAI           \\
BERT-ViT & <1B & Local & Liang et al. \\
TMPT & <1B & Local & Liang et al. \\
LLaVA-1.5-7B & 7B & Local & Microsoft \\
Qwen2.5-VL-7B & 7B & Local & Alibaba          \\
MiMo-VL-7B & 7B & Local & Xiaomi          \\
GPT4-1 & Unknown  & API   & OpenAI           \\
\bottomrule
\end{tabular}
\end{table}

\section{Prompt Templates}
\label{sec:appendix-Prompt}
To support the reproducibility and interpretability of our methodology and experimental settings, we provide the comprehensive set of prompt templates used in this work. Specifically, Table \ref{tab:prompt-Persona} details the prompts for User Persona Distillation; Table \ref{tab:prompt-Intent} presents the two-stage prompt protocol for Image Intent Extraction; and Table \ref{tab:prompt-stance} outlines the prompt employed for the baseline models.

\begin{table*}[t]
\centering
\small
\begin{CJK*}{UTF8}{gbsn}
\begin{tcolorbox}[enhanced, sharp corners, boxrule=0.4pt, colback=gray!3, colframe=black, width=\textwidth, title=B.1~User Persona Distillation]

\medskip
\textbf{System Prompt}:\par
You are an expert in computational psychology and linguistic analysis. Your task is to analyze a user's Big Five personality traits (Openness, Conscientiousness, Extraversion, Agreeableness, and Neuroticism) based on their historical social media activity.

\medskip
\textbf{User Prompt}:\par
\#\#\# Task Description:\par
Analyze the provided user profiles, including their posts and comments. For each of the five traits, provide a qualitative description of how the user exhibits this trait and a quantitative score on a scale of 1-5 (1: Very Low, 5: Very High).\par
\medskip
\#\#\# Big Five Traits Definition:\par
1. Openness to Experience: Curiosity, originality, and openness to new ideas.\par
2. Conscientiousness: Organization, responsibility, and goal-oriented behavior.\par
3. Extraversion: Sociability, assertiveness, and emotional expressiveness.\par
4. Agreeableness: Trust, altruism, kindness, and affection.\par
5. Neuroticism: Emotional instability, anxiety, and irritability.\par
\medskip
\#\#\# Input Data:\par
User ID: \{\{user id\}\}\par
Recent Posts: \par
\{\{post list\}\}\par
Recent Comments:\par
\{\{comment list\}\}\par
\medskip
Output:
\end{tcolorbox}
\end{CJK*}
\caption{Prompt template for User Persona Distillation. The model outputs a summary of Big Five personality traits for users.}
\label{tab:prompt-Persona}
\end{table*}

\begin{table*}[t]
\centering
\small
\begin{CJK*}{UTF8}{gbsn}
\begin{tcolorbox}[enhanced, sharp corners, boxrule=0.4pt, colback=gray!3, colframe=black, width=\textwidth, title=B.2~Image Intent Extraction]

\medskip
\textbf{Step 1}:\par
\medskip
\textbf{System Prompt}:\par
You are an expert image analyzer.

\medskip
\textbf{User Prompt}:\par
\#\#\# Task Description:\par
You will be given an image. Your task is to produce a detailed and structured description of the image. Please follow the required steps:\par
1. **Overall Scene Description**  \par
   Describe the setting, environment, and general context of the image.\par
2. **Key Objects and Entities**  \par
   Identify and describe the notable objects, people, or elements in the image, including their appearance, attributes, and spatial relationships.\par
3. **Fine-Grained Visual Details**  \par
   Highlight colors, expressions, textures, gestures, body language, or any small details that may convey meaning.\par
4. **Style and Emotional Tone (if applicable)**  \par
   Describe the mood, atmosphere, artistic style, or emotional impression conveyed by the image.\par
\medskip
\#\#\# Input: <image>\par
\medskip
\#\#\# Output Requirment:\par
Make sure your description is precise, neutral, and avoids assumptions unless clearly supported by visual evidence.\par
Directly output the description text**, starting with **`Image Description:`** followed by your concise answer.\par
\medskip
Output:

\medskip

\textbf{Step 2}:\par
\medskip
\textbf{System Prompt}:\par
You are an expert in multimodal conversation analysis.\par
\medskip
\textbf{User Prompt}:\par
\#\#\# Task Description:\par
Determine the **intention or communicative purpose** of the user who posted the image in the conversation.  \par
Consider the social and semantic context of the conversation, as well as the content of the image.\par
\medskip
You are given:\par
1. A **description of the image**, which has been generated by a visual-language model.\par
2. A **conversation history** between multiple participants.  \par
   - In this history, the placeholder "<image>" marks where the image was posted by the user.\par
\medskip
If the conversation contains multiple images, carefully identify the **index or order** of the images (e.g., <image1>, <image2>, <image3>...) and determine where the **current target image** appears in the conversation before analyzing it.\par
\medskip
\#\#\# Requirment:\par
Follow these steps carefully:\par
1. Review the conversation history and understand the flow of dialogue.  \par
2. Interpret the meaning of the image based on the provided description.  \par
3. Analyze how the image contributes to the conversation — for example, whether it expresses emotion, humor, evidence, sarcasm, agreement, mockery, or provides visual information.  \par
4. Infer the **most likely intention** of the user when posting the image.\par
5. Directly output the intent text**, starting with **`Intent:`** followed by your concise answer.\par
\medskip
\#\#\# Image Description:\par
The image captures a bustling urban scene at night, likely Times Square in New York City, characterized by its iconic billboards and vibrant advertisements. The foreground features ...\par
\medskip
\#\#\# Conversation History:\par
User youyouxue: We just drove from San Francisco to New York City in a Tesla Model X without using a single drop of gas! <image1>\par
User salec1:I did a shorter version of the same thing (Miami to NYC) last month. It's all about ... \par
User KickAClay:I always wondered if any Tesla, after inputting your route ... \par
User timdorr:Yes, it does exactly this. Set your destination and it will find Superchargers along your route, tell you how long to ... \par
User chupacabrando:How much longer does it take ...\par
...\par
\medskip
Output:
\end{tcolorbox}
\end{CJK*}
\caption{Prompt template for Image Intent Extraction. The model outputs a detailed and structured description of the image.}
\label{tab:prompt-Intent}
\end{table*}

\begin{table*}[t]
\centering
\small
\begin{CJK*}{UTF8}{gbsn}
\begin{tcolorbox}[enhanced, sharp corners, boxrule=0.4pt, colback=gray!3, colframe=black, width=\textwidth, title=B.3~Zero-shot Stance Detection]

\medskip
\textbf{System Prompt}:\par
You are an expert linguist specializing in conversational analysis and social media stance detection.\par

\medskip
\textbf{User Prompt}:\par
\#\#\# Task Description:\par
Analyze the provided conversation flow and determine the stance of the current speaker toward the specific Target listed below. The attached image(s) correspond to the sequential labels (e.g., <image1>, <image2>) mentioned in the text.\par
\medskip
\#\#\# Target:\par
Tesla\par
\medskip
\#\#\# Stance Classes:\par
['favor','against','none']\par
\medskip
\#\#\# Conversation History:\par
User youyouxue: We just drove from San Francisco to New York City in a Tesla Model X without using a single drop of gas! <image1>\par
User salec1:I did a shorter version of the same thing (Miami to NYC) last month. It's all about ... \par
User KickAClay:I always wondered if any Tesla, after inputting your route ... \par
User timdorr:Yes, it does exactly this. Set your destination and it will find Superchargers along your route, tell you how long to ... \par
User chupacabrando:How much longer does it take ...\par
...
\medskip
\#\#\# Current Speaker's Reddit Profile:\par
Active Communities:\par
--- Historical Posts ---\par
...\par
--- Historical Comments ---\par
...\par
\medskip
\#\#\# Output Format (JSON):\par
{\"stance\": "favor" | "against" | "none"}\par
\medskip
Output:
\end{tcolorbox}
\end{CJK*}
\caption{Prompt template for Multimodal Conversational Stance Detection. The model outputs a stance class for the current instance.}
\label{tab:prompt-stance}
\end{table*}

\end{document}